\documentclass[10pt,twocolumn,letterpaper]{article}

\usepackage{wacv}
\usepackage{times}
\usepackage{epsfig}
\usepackage{graphicx}
\usepackage{amsmath}
\usepackage{amssymb}
\usepackage{subcaption}

\usepackage{caption} 
\def\wacvPaperID{0008} 

\wacvfinalcopy 

\ifwacvfinal
\fi

\ifwacvfinal
\usepackage[breaklinks=true,bookmarks=false]{hyperref}
\else
\usepackage[pagebackref=true,breaklinks=true,colorlinks,bookmarks=false]{hyperref}
\fi

\ifwacvfinal
\else
\fi

\begin{document}


\title{On Salience-Sensitive Sign Classification in Autonomous Vehicle Path Planning: Experimental Explorations with a Novel Dataset}

\author{
Ross Greer\\
{\tt\small regreer@ucsd.edu}

\and
Jason Isa\\
{\tt\small jisa@ucsd.edu}

\and 
Nachiket Deo\\
{\tt\small ndeo@ucsd.edu}

\and 
Akshay Rangesh \\
{\tt\small arangesh@ucsd.edu}

\and
Mohan M. Trivedi\\
{\tt\small mtrivedi@ucsd.edu}\\
\\
University of California, San Diego\\
La Jolla, CA\\
}

\maketitle
\pagenumbering{gobble}
\begin{abstract}
Safe path planning in autonomous driving is a complex task due to the interplay of static scene elements and uncertain surrounding agents. While all static scene elements are a source of information, there is asymmetric importance to the information available to the ego vehicle. We present a dataset with a novel feature, sign salience, defined to indicate whether a sign is distinctly informative to the goals of the ego vehicle with regards to traffic regulations. Using convolutional networks on cropped signs, in tandem with experimental augmentation by road type, image coordinates, and planned maneuver, we predict the sign salience property with 76\% accuracy, finding the best improvement using information on vehicle maneuver with sign images. 
\end{abstract}

\section{Introduction}

Autonomous vehicles need to share the road with multiple decision making agents, each with their own goals and different directions of motion. Traffic signs play an important role in regulating the motion of all agents on the road. They are easy to notice and provide safety critical information
in an intuitive and succinct manner to drivers and pedestrians. Traffic sign detection and recognition has thus received significant attention in recent research on autonomous driving and advanced driver assistance systems (ADAS). Several datasets have been released, with bounding boxes (\cite{Stallkamp2012} \cite{ertler2020mapillary} \cite{mogelmose2014traffic} \cite{Zhe_2016_CVPR} \cite{temel2019traffic} \cite{larsson2011using} \cite{huang} \cite{shakhuro2016russian}), pixel level masks (\cite{Zhe_2016_CVPR}), as well as fine-grained category labels for traffic signs (\cite{Stallkamp2012}  \cite{ertler2020mapillary} \cite{mogelmose2014traffic} \cite{Zhe_2016_CVPR} \cite{huang} \cite{shakhuro2016russian}). These in turn have allowed researchers to leverage modern CNN based detectors and classifiers for traffic sign detection and recognition (\cite{zhang2020cascaded} \cite{cao2021traffic} \cite{arcos2018deep}).

\begin{figure*}
    \centering
    \includegraphics[width=\textwidth]{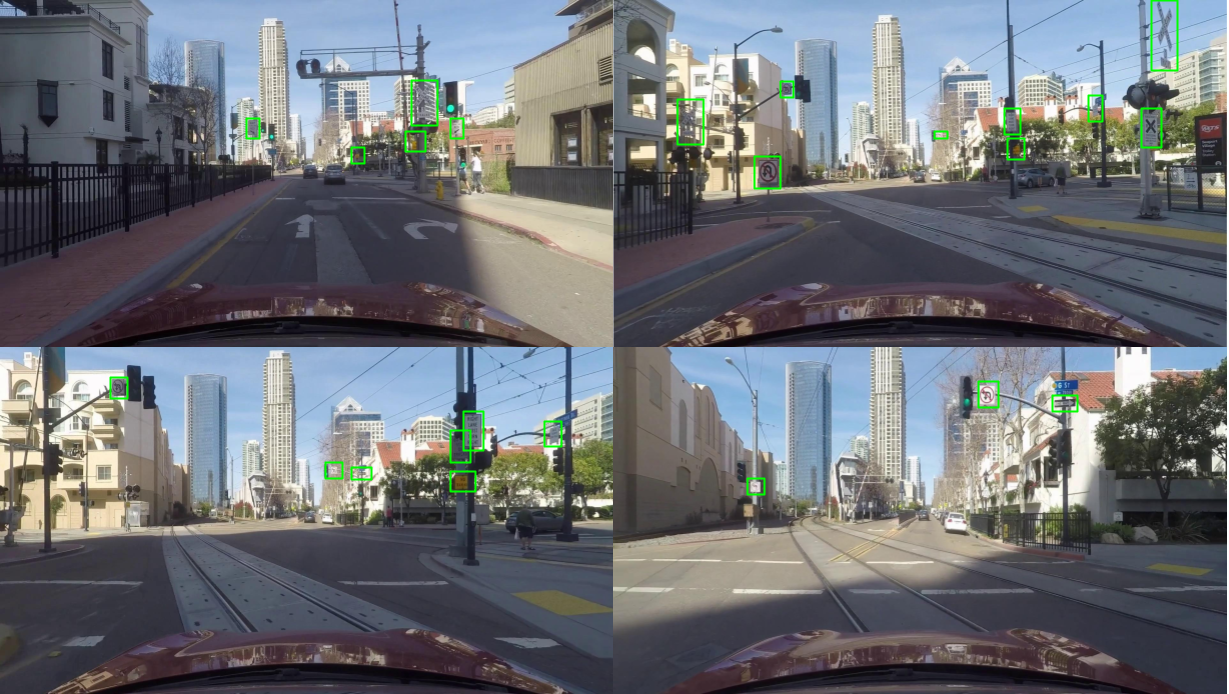}
    \caption{In this montage, a vehicle drives through a complex scene containing a heavy amount of signs of varying salience to the intended path. There are signs in the field of view which instruct the lanes to the left and right of the ego lane, as well as the cross-traffic at the intersection. While informative about the possible paths of other agents, these signs do not provide direction to the vehicle in proposing its own path through the intersection.}
    \label{fig:many_signs}
\end{figure*}

\begin{figure}
    \centering
    \includegraphics[width=.5\textwidth]{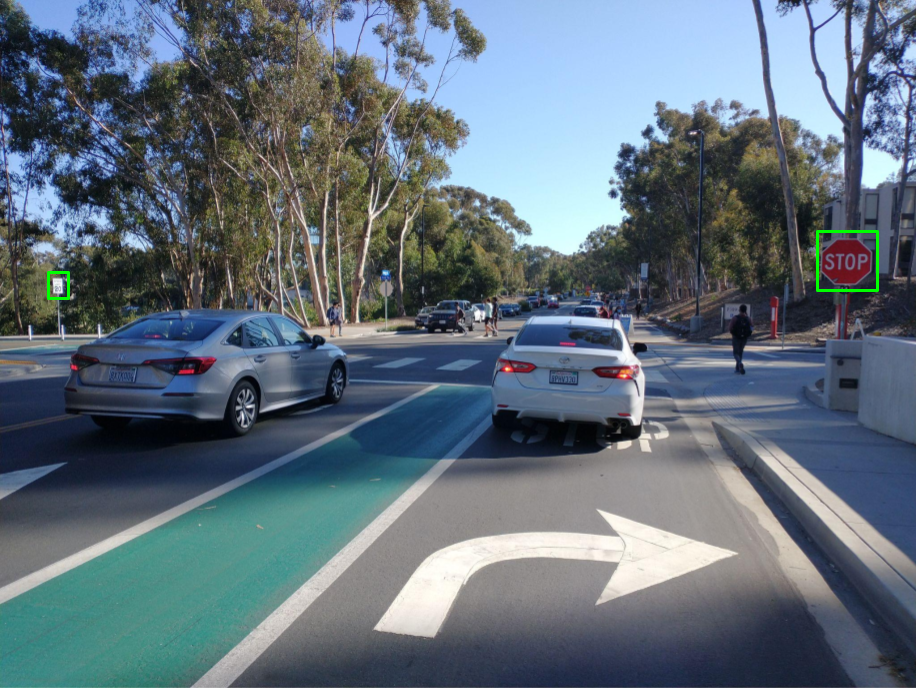}
    \caption{In a model designed to detect and classify signs for safe autonomous driving, being aware of the stop sign is much more important than the cross-traffic speed limit. A model should be able to weigh missed detections accordingly, as made possible with the sign salience property. By the proposed definition, the stop sign is salient while the speed limit sign is not.}
    \label{fig:stop_geisel}
\end{figure}

\begin{figure}
    \centering
    \includegraphics[width=.5\textwidth]{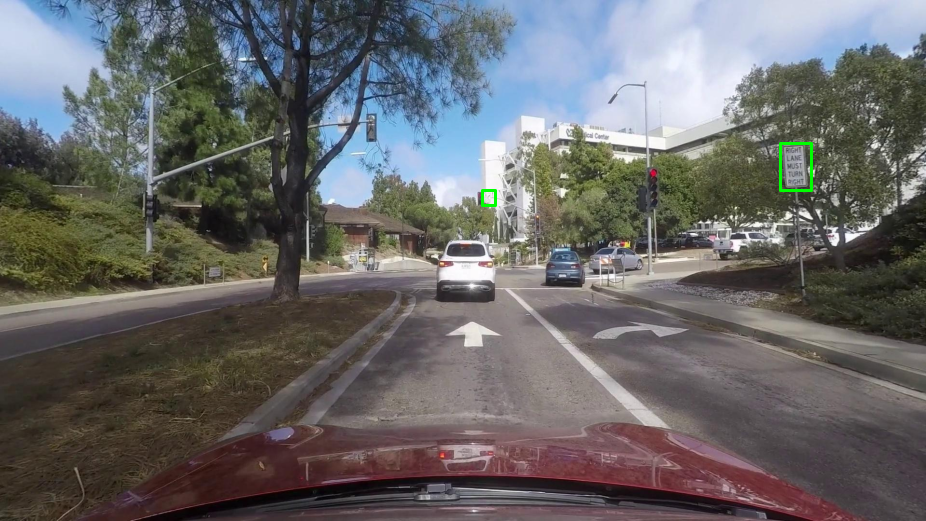}
    \caption{In this scenario, two possible sign detections are made, but while the detection on the right is easier, it provides no value to the ego vehicle in understanding allowable maneuvers in the upcoming intersection. Detecting this sign is less critical, but existing traffic sign datasets do not contain features with this information. Additionally, were the autonomous vehicle to mistakenly associate this ``Must Turn Right" sign as salient to its lane, it would make an illegal maneuver by following its instruction. Only the white regulatory sign located across the intersection is salient.}
    \label{fig:turn_right}
\end{figure}

\begin{figure}
    \centering
    \includegraphics[width=.5\textwidth]{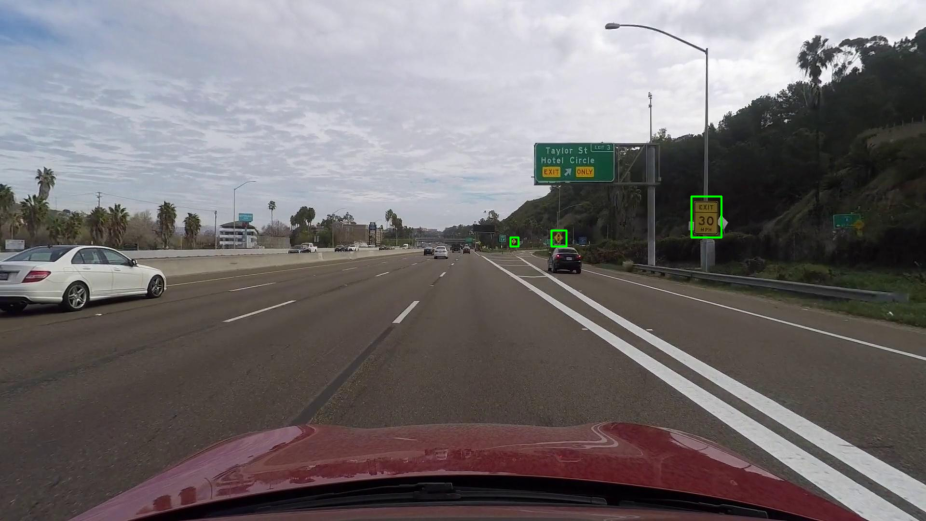}
    \caption{While most speed limit signs apply to all lanes in the direction of travel, this exit speed limit applies only to the lane to the right of the ego vehicle. An autonomous vehicle must have the ability to determine whether this sign is salient to its lane, and select its speed accordingly. The detected signs in this scene are not salient.}
    \label{fig:exit_speed}
\end{figure}

While detection and recognition of traffic signs are important tasks, they aren't sufficient to inform an autonomous vehicle how to operate in a traffic scene. Crucially, an autonomous vehicle needs the ability to determine whether a traffic sign is \textit{salient} or applicable to its planned path. This is a challenging task due to several factors:
 
\begin{itemize}

\item \textbf{Scene complexity}: City streets are complex environments. Consider the montage shown in Figure \ref{fig:many_signs}; in addition to being a visual maze on its own (between lane flows, non-perpendicular intersections, and train tracks) the scene contains excessive sign  information,  where  the  controller  must  know  which signs are meant to inform its own lane and not follow
the signs intended for others.

\item \textbf{Asymmetric importance of scene elements}: While there is information available in every pixel visible to a vehicle, autonomous or manually-driven, certain portions of a given scene are more important in path planning. As a motivating example, being aware of a speed limit sign directed at cross traffic certainly informs a driver of expected behavior of other vehicles in the scene, but is less relevant to the driver's plans than an imminent stop sign, as illustrated in Figure \ref{fig:stop_geisel}.

\item \textbf{Extraneous traffic signs}: In other cases, the sign which is easiest to detect may not necessarily be instructive for the ego vehicle. Consider Figures \ref{fig:turn_right} and \ref{fig:exit_speed}, where the closest sign on the right, while in a typically informative location, actually provides information to a different lane than the ego vehicle, and following such instructions may prove dangerous and unexpected to surrounding drivers. 

\item \textbf{Non-local context cues}: The context cues to determine whether a traffic sign applies to the ego-vehicle can often be non-local to the traffic sign itself. These non-local cues could ego-vehicle's lane, its planned route, and in some cases (such as yield signs) even the locations of surrounding agents.
\end{itemize}

This paper represents a first step towards traffic sign salience recognition. We define a sign to be \textit{salient} if the visible sign provides an instruction intended for the ego vehicle location before the next decision point, independent of the actions of other agents and instructions provided to other lanes. To facilitate further research on traffic sign salience recognition, we present the LAVA traffic sign dataset with traffic sign bounding boxes, fine-grained traffic sign category labels, as well as binary labels indicating traffic sign salience. Additionally we provide auxiliary meta-data for each scene including roadway type, and the next planned maneuver for the ego-vehicle. Finally, we present analysis on traffic sign salience recognition using a CNN based classifier that takes into account the appearance of traffic signs, their locations in a given scene, the roadway type (e.g. highway, intersection, on-ramp, school-zone etc.), and the planned route of the ego-vehicle.

\section{Related Research}

\subsection{Sign Detection}

The task of monocular sign detection is well-established and well-addressed in the field, with prime evidence in the nearly-perfect precision-recall curves associated with the German Traffic Sign Detection Benchmark public results \cite{Stallkamp2012}. Recent approaches of significant performance across multiple datasets include:
\begin{itemize}
    \item A cascaded R-CNN with multiscale attention \cite{zhang2020cascaded}, with data augmentation to balance class prevalence of commonly-missed small signs. This method is designed to address false detections due to illumination variation and bad weather.
    \item  A sparse R-CNN with residual connections in the ResNest backbone and a self-attention mechanism \cite{cao2021traffic}, designed to be robust to foggy, frosty, and snowy images. 
\end{itemize}

In our work, we assume prior knowledge of the detected sign location, as would be given using any of the above methods. 

\subsection{Sign Classification}

While sign classification in itself is not necessary in our model of sign importance, the problem has been addressed to high levels of accuracy in \cite{arcos2018deep}, the top performer on the German Traffic Sign Recognition Benchmark, using a CNN with three spatial transformers. Our work predicts sign salience using standard convolutional filter features extracted from the cropped sign image, but ongoing SOA approaches to sign recognition can provide improved backbone features to salience classification, since a sign's appearance can certainly affect its relevance (e.g. a stop sign detected while the ego vehicle is on a freeway is likely meant for off-freeway traffic and is therefore irrelevant). Further, sign classification can be combined with sign salience for downstream control, such that the control module can understand first which signs are important, and second what expected behaviors those important signs are indicating. 

\subsection{Traffic Sign Datasets}

The traffic sign datasets listed in Table \ref{datasets} facilitate research in the above tasks of traffic sign detection and classification. In this table, we highlight the relative size of these datasets, as well as any unique annotated features beyond the traffic sign class and bounding box image coordinates. 

\begin{table*}
\begin{center}
 \small
\begin{tabular}{||c c c c||} 
 \hline
 Dataset & Number of Images & Country & Features \\ 
 \hline\hline

 German Traffic Sign Detection Benchmark \cite{Stallkamp2012} & 50,000+ & Germany &  \\ 
 \hline
 Mapillary \cite{ertler2020mapillary} & 100,000 & World & \\
 \hline
 LISA Traffic Sign Dataset \cite{mogelmose2014traffic} & 7,855 & USA & occlusion, on-side-road \\
 \hline
 Tsinghua-Tencent 100K \cite{Zhe_2016_CVPR} & 100,000 & China & pixel-level mask \\
 \hline
 CURE-TSD \cite{temel2019traffic} & 1.7M & Belgium & \\  
 \hline
  LiU Traffic Signs Dataset \cite{larsson2011using} & 3,488 & Sweden & occlusion \\  
 \hline
  Chinese Traffic Sign Database \cite{huang} & 6,164 & China &  \\  
 \hline
  Russian Traffic Sign Images Dataset \cite{shakhuro2016russian} & 104,358 & Russia & \\ 
 \hline
 LISA Amazon-MLSL Vehicle Attributes Dataset \cite{kulkarni_rangesh_buck_feltracco_trivedi_deo_greer_sarraf_sathyanarayana_2021} & 14,112 & USA & 10s video context, occlusion, \textbf{salience} \\ 
 \hline
\end{tabular}
\caption{Comparison of traffic sign datasets. All datasets contain at least images, class labels, and information about the image location and size of the bounding box for the region of interest, in addition to any unique features listed. The LAVA dataset (bottom row) is notable for its inclusion of 10 second video context, and importantly is the only dataset which contains a binary label indicating sign salience.}
\label{datasets}

\end{center}
\end{table*}

\normalsize

\subsection{Traffic Signs in Planning and Control}

Most current consumer-market vehicles offer little path planning for traffic regulations apart from maintaining lane, maintaining reasonable speed, and avoiding collision, hence advisory restricting the use of these features to freeways-only. Autonomous vehicles are expected to come with minimum safety guarantees, but the verification and explainability of such systems is difficult to address with common end-to-end learning approaches. Integration of algorithms which address traffic regulations can provide the deterministic and explainable action qualities important to public acceptance and safety. As explained by Fulton et al. \cite{fulton2020formal}, ``Autonomous systems that rely on formally constrained RL for safety must correctly map from sensory inputs into the state space in which safety specifications are stated. I.e., the system must correctly couple visual inputs to symbolic states." A recent approach to address this explainability, Cultrera et al. \cite{Cultrera_2020_CVPR_Workshops} use end-to-end visual attention model which allows identification of what parts of the image the model has deemed most important. The specific importance of regulatory scene understanding has proven useful in trajectory prediction by Greer et al. \cite{greer2021trajectory}, using a weighted lane-heading loss to ascribe importance to lane-following. Learned attention to the static scene has been demonstrated effective by Messaoud et al. \cite{messaoud2020trajectory}, showing that end-to-end approaches to trajectory prediction which take in only agent motion are missing valuable information from the regulations of the static scene.  

As an example of a recent model which acknowledges the importance of sign-adherence capabilities, \cite{guo2020vision} use an end-to-end learning approach  to control using navigational commands, but note a shortcoming: ``Traffic rules such as traffic lights, and stop signs are ignored in the dataset, therefore, our trained model will not be able to follow traffic lights or stop at stop signs." Some regulations can be addressed by algorithms (rule-based or learned) which are tailored to specific signs. Alves et al. \cite{alves2021double} explore planning under traffic sign regulations by modelling and implementing three Road Junction rules involving UK stop and give-way signs. 

While this work is the first to ascribe and predict sign salience, Guo et al. \cite{10.1145/3474085.3475362} create a dataset and learning architecture to promote descriptive understanding of signs beyond common detection and recognition. Their model is intended to output a semantic, verbal description which connects the texts and symbols on a sign. In contrast, our method does not seek to understand the semantic meaning of the sign, but rather whether the sign is important to the attention of the vehicle. These features are clearly informative to one another, but while their output is intended to influence navigational decisions, our output is better fit for loss-weighting schemes in safety-critical detections and recognitions.

\subsection{Road Object Salience}

Identifying salient objects has been explored in connection with driver behavior analysis; knowing where a driver is looking can be a predictor of scene salience, but alternatively, knowing salient objects prior to observing gaze can inform an intelligent vehicle of possible gaps in the driver's attention. Dua et al. \cite{dua2020dgaze} create the DGAZE dataset to map driver gaze in scene images, connecting gaze to driver focus and attention, topics extensively studied in connection to safe, highly-automated driving in the recent survey by Kotseruba and Tsotsos \cite{kotseruba2021behavioral}. Lateef et al. \cite{lateef2021saliency} use a GAN to predict important objects in driving scenes, with data from existing driving datasets labeled using a salience mechanism which weights object classes from the semantic segmentation of the scene, building a Visual Attention Driving Database. Su et al. \cite{su2021exploring} show that salience is a property which can transfer from non-driving-related tasks to driving tasks, learning attention on CityScapes from standard salient object detection (SOD) datasets. Pal et al. \cite{pal2020looking} show that combining static scene information with driver gaze information in their SAGE-Net can propose important regions of attention. 

Li et al. \cite{li2020make} define the task of risk object identification, under the hypothesis that objects influencing drivers' behavior are risky. Though their work is intended to cover a more general scope of objects than traffic signs, signs that we determine to be salient do hold a similar property; that is, were the sign not present, it is possible that the driver's behavior would change. However, there are cases where the sign is intended to create an awareness of surrounding scene elements, in which case the sign would still be salient by our definition, but not necessarily a risk object. Zhang et al. \cite{zhang2020interaction} agree to the importance of salience analysis, stating ``A vehicle driving along the road is surrounded by many objects, but only a small subset of them influence the driver’s decisions and actions. Learning to estimate the importance of each object on the driver’s real-time decisionmaking may help better understand human driving behavior and lead to more reliable autonomous driving systems." Their work builds this estimation using interaction graphs which allow for the importance of scene elements to change depending on interactions observed between other scene elements (without involvement of the ego-vehicle). Our work is completely driver-centric; that is, we consider here signs which address the ego vehicle independent of the actions of other scene agents. 

\section{LAVA Dataset for Salient-Sensitive Traffic Signs}

The LISA Amazon-MLSL Vehicle Attributes (LAVA) Dataset \cite{kulkarni_rangesh_buck_feltracco_trivedi_deo_greer_sarraf_sathyanarayana_2021} includes a collection of traffic signs bounded and labeled in images taken from a front-facing camera, including 10 second video clips for full scene and trajectory context, accompanied by INS data. The data has been collected from the greater San Diego area, curated in a manner which includes a diversity of road types, traffic conditions, weather, and lighting. The traffic signs are categorized as stop, yield, do not enter, wrong way, school zone, railroad, red and white regulatory, white regulatory, construction and maintenance, warning, no turn, one way, no turn on red, do not pass, speed limit, guide, service and recreation, and undefined. The frequency of the sign types are described in Table \ref{signtypes}. Signs are given a tag if electric (0.18\%) or occluded (11.86\%), and each sign is assigned an \textit{is\_salient} property with respect to the position of the ego vehicle (66.42\%). For experimentation, we divide the 14,112 samples into 11,289 training instance, 1,411 validation instances, and 1,412 test instances, ensuring no scene is divided between sets. 

\begin{table}
\begin{center}
\begin{tabular}{||c c||} 
 \hline
 Sign Type & Frequency \\ 
 \hline\hline
 Stop & 725 \\
 \hline
 Yield & 72  \\
 \hline
 Do Not Enter & 134 \\
 \hline
 Wrong Way & 51 \\  
 \hline
 School Zone & 172 \\
 \hline
 Railroad & 7 \\
 \hline
 Red \& White Regulatory & 710  \\
 \hline
 White Regulatory & 3,048 \\
 \hline
 Construction \& Maintenance & 773 \\  
 \hline
 Warning & 2,364 \\
 \hline
 No Turn & 419 \\
 \hline
 No Turn on Red & 224 \\
 \hline
 One Way & 109  \\
 \hline
 Do Not Pass & 9 \\
 \hline
 Speed Limit & 563 \\  
 \hline
 Guide & 249 \\
 \hline
 Service \& Recreation & 2 \\
 \hline
 Undefined & 833  \\
 \hline
\end{tabular}
\caption{Sign type frequencies in the LAVA dataset. The data is non-uniformly distributed, reflecting an approximate real-world distribution of signs within the region of collection.}
\label{signtypes}
\end{center}
\end{table}

\subsection{Automatic Road Type Classification}

Reducing video and image data to mid-level semantic drive information has been shown to be important in understanding naturalistic drive data \cite{6910285}. Similarly, we posit that information such as road scene and drive maneuver may contain important contextual information related to sign salience. Beyond traffic sign information explained above, we further classify each image scenes found in the LAVA dataset into 12 possible classes: highway, city street, residential, roundabout, intersection, construction zone, tunnel, freeway entrance, freeway exit, and unknown. This classification is performed automatically as follows:

\begin{itemize}
    \item \textit{Road Type by Global Coordinates}: LAVA sensor data includes latitude and longitude coordinates associated with each frame. Using Nominatim's reverse geocoding \cite{clemens2015geocoding}, we find the name and OpenStreetMap [OSM] ID of the current street. OSM provides categories of motorway, primary, secondary, tertiary, trunk, residential, roundabout, and pedestrian. We map primary, secondary, tertiary, and trunk to city street, motorway to highway, residential to residential, roundabout to roundabout, and pedestrian to parking lot. This excludes the classes of intersection, construction zone, tunnel, school zone, freeway entrance, and freeway exit.
    \item \textit{Road Type by Object Detection}: We use CenterNet \cite{zhou2019objects} trained on NuScenes \cite{caesar2020nuscenes} for detecting traffic signs, lights and traffic cones in the scene. If an image is detected to contain two or more traffic cones, it is classified as construction zone. Similarly, if one or more stop sign is detected, or three or more traffic lights are detected, the image is classified as an intersection. 
    \item \textit{Road Type by Class Change}: Using a contextual 10 second clip, if the frame class begins as highway and transitions to city street, the images of the clip are re-labeled as freeway entrance. Similarly, in the reverse case, the images are re-labeled as freeway exit.
\end{itemize}

The frequency with which signs are found on a particular road type are summarized in Table \ref{roadtypes}. 

\begin{table}
\begin{center}
\begin{tabular}{||c c||} 
 \hline
 Road Type & Frequency \\ 
 \hline\hline
 Highway & 1,788 \\
 \hline
 City Street & 8,285  \\
 \hline
 Residential & 1,243 \\
 \hline
 Roundabout & 17 \\  
 \hline
 Intersection & 972 \\
 \hline
 Construction Zone & 300  \\
 \hline
 Freeway Entrance & 228 \\  
 \hline
 Freeway Exit & 207 \\
 \hline
 Unknown & 1,072 \\
 \hline
\end{tabular}
\caption{Road type frequencies per sign in the LAVA dataset.}
\label{roadtypes}
\end{center}
\end{table}

\subsection{Maneuver Classification}

For each frame in the LAVA dataset, we analyze the following 10 seconds of vehicle speed and yaw data for rule-based classification of the intended driving maneuver as Forward, Stop, Turn Left After Stopping, Turn Right After Stopping, Turn Left, and Turn Right. The frequency of maneuvers are described in Table \ref{maneuvers}. 

\begin{table}
\begin{center}
\begin{tabular}{||c c||} 
 \hline
 Maneuver & Frequency \\ 
 \hline\hline
 Forward & 8,593 \\
 \hline
 Stop & 4,535 \\
 \hline
 Turn Left After Stopping & 18 \\
 \hline
 Turn Right After Stopping & 41 \\  
 \hline
 Turn Left & 476 \\
 \hline
 Turn Right & 449 \\
 \hline
\end{tabular}
\caption{Maneuver frequencies per sign in the LAVA dataset.}
\label{maneuvers}
\end{center}
\end{table}

\section{Sign Salience Prediction}
\subsection{On-Right Classifier Baseline}

While a random classifier would give an expected accuracy of 50\%, we consider a reasonable trivial classifier which is better-grounded in traffic sign priors. This classifier assigns salience to signs which are located on or to the right of center, and non-salience to signs which are located left of center. This is consistent with typical drive-on-the-right traffic flow in the US, and should be adjusted for countries which drive on the left when comparing across datasets.

\subsection{ResNet50 Model}

We begin from the hypothesis that visual information can be used as a preliminary indicator of sign salience. The convolutional model uses the ResNet50 \cite{he2016deep} convolutional architecture typically found in sign recognition. It takes a cropped sign region as input, and outputs a binary label for salience. We use an Adam optimizer with an initial learning rate of $10^{-3}$ and a batch size of 64.

\subsection{Road Type Augmentation}

To improve performance beyond the ResNet50 model, we consider the effects of road type on expected sign salience. Certain road types are less likely to see particular relevant signs; for example, a stop sign is unlikely to appear on a freeway, and a 65 MPH speed limit is unlikely to appear in a school zone. For this reason, if the features of such a sign are found in the convolutional layers, it is likely that the detected sign belongs to a different road or lane than that of the ego vehicle (perhaps past an off-ramp or overpass). 

To test this hypothesis, we augment the model by appending a one-hot encoded vector representing the perceived road type to the flattened convolutional output prior to the fully-connected layer. We then add a ReLU activation, followed by another fully-connected layer, another ReLU, and a final fully-connected layer before the softmax binary output activation. Until the last binary activation, we maintain 2,048 nodes at each fully-connected layer.

\subsection{Image Coordinate Augmentation}

\begin{figure}
    \centering
         \begin{subfigure}[b]{.5\textwidth}
         \centering
         \includegraphics[width=\textwidth]{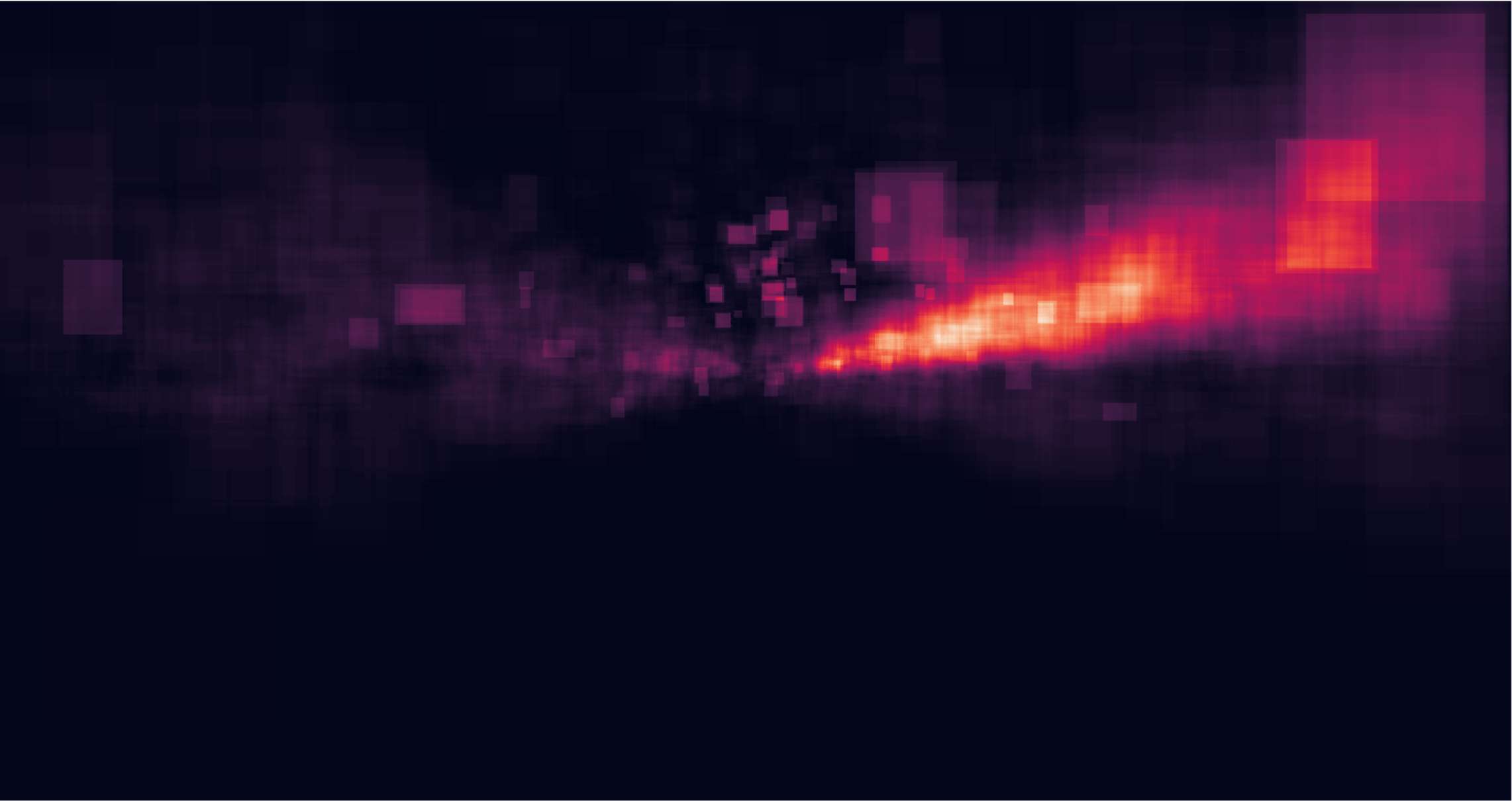}
         \caption{}
        \end{subfigure}
         \begin{subfigure}[b]{.5\textwidth}
         \centering
         \includegraphics[width=\textwidth]{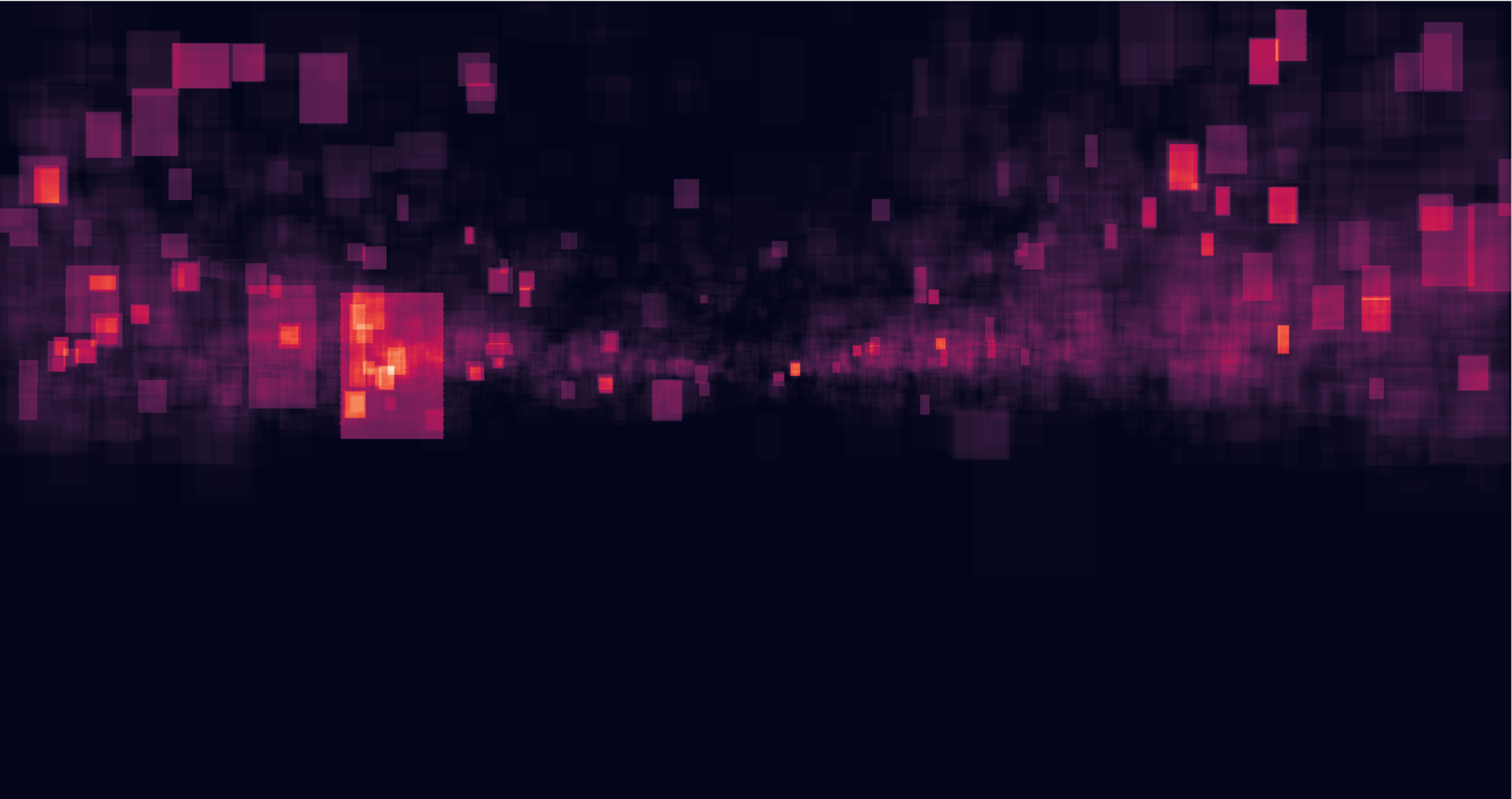}
         \caption{}
        \end{subfigure}
    \caption{Heatmaps illustrating frequency with which a pixel is occupied by (a) a salient sign or (b) a non-salient sign.}
    \label{fig:isfront}
\end{figure}

Another feature which may improve model performance is the information contained in the pixel size and location of the detected bounding box within the scene image. In general, salient signs are found to the right of center and above the ego vehicle, as illustrated in Figure \ref{fig:isfront}, and there is a relationship between the size of the sign and its location which can provide information about the 3D depth of the sign. This depth information provides further context to the model about where the sign may be located relative to the ego vehicle. 

We augment the model by appending the top-left coordinate $(x, y)$ of the bounding box (normalized to the image width and height), the bounding box width $w$, and bounding box height $h$ to the flattened convolutional output prior to the fully-connected layer. Similar to the above road type augmentation, we then add a ReLU activation, followed by another fully-connected layer, another ReLU, and a final fully-connected layer before the softmax binary output activation. Until the last binary activation, we maintain 2,048 nodes at each fully-connected layer. 

\begin{figure*}
    \centering
         \begin{subfigure}[b]{.49\textwidth}
         \centering
         \includegraphics[width=\textwidth]{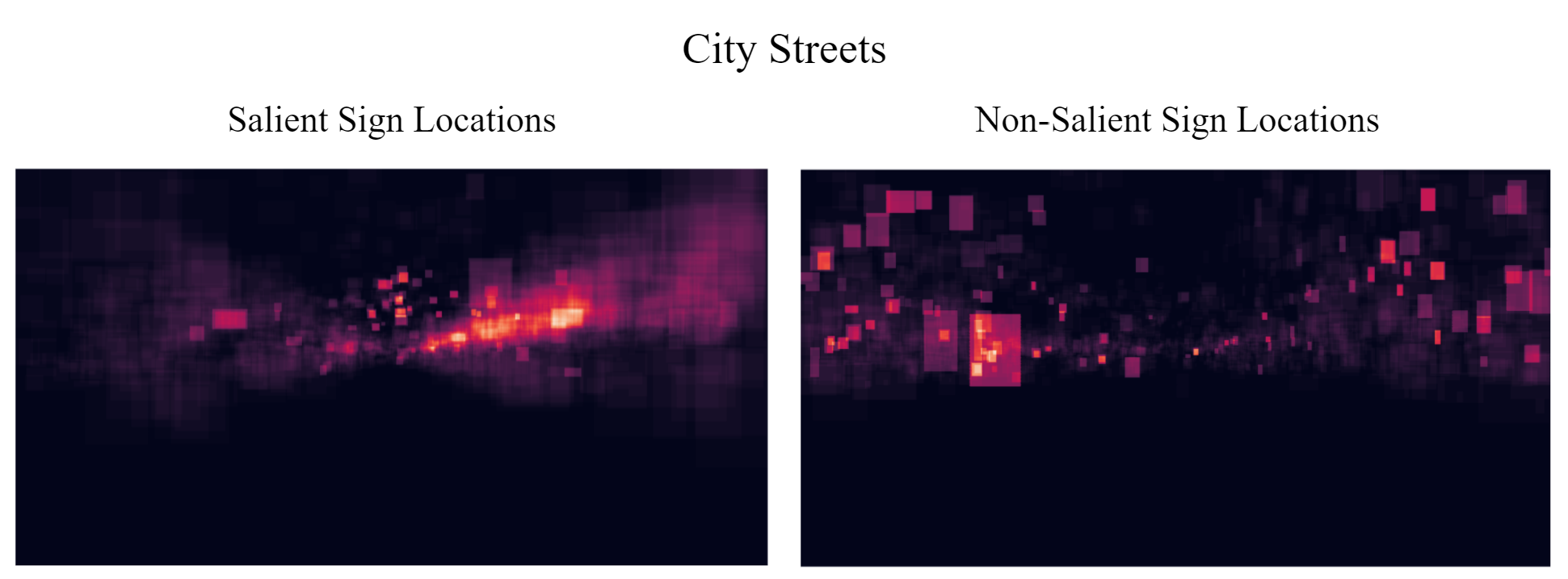}
         \caption{}
        \end{subfigure}
         \begin{subfigure}[b]{.49\textwidth}
         \centering
         \includegraphics[width=\textwidth]{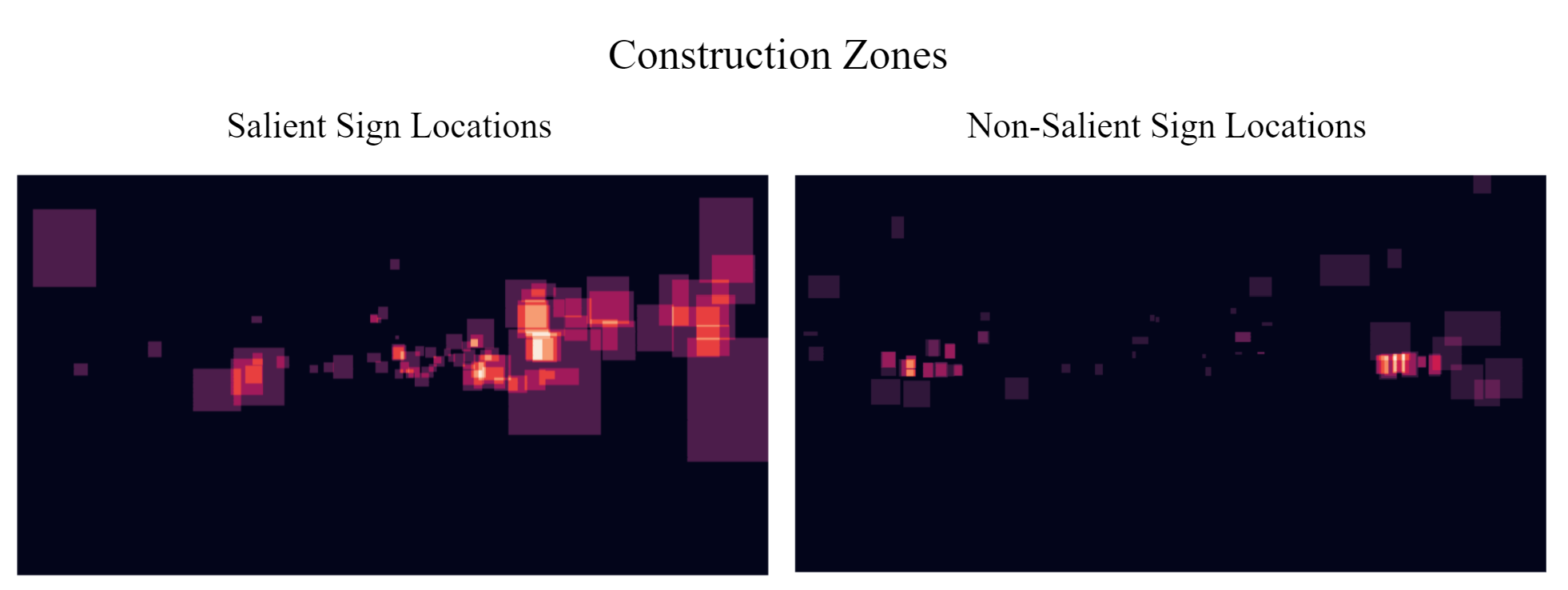}
         \caption{}
        \end{subfigure}

         \begin{subfigure}[b]{.49\textwidth}
         \centering
         \includegraphics[width=\textwidth]{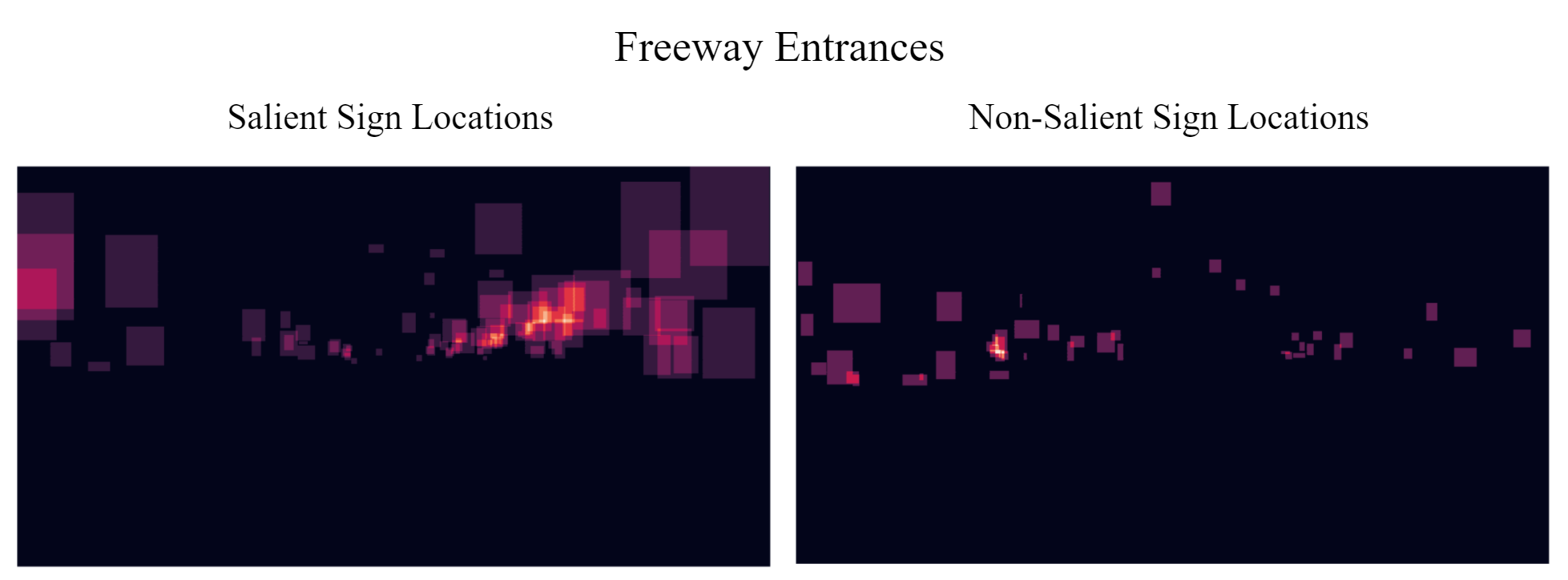}
         \caption{}
        \end{subfigure}
         \begin{subfigure}[b]{.49\textwidth}
         \centering
         \includegraphics[width=\textwidth]{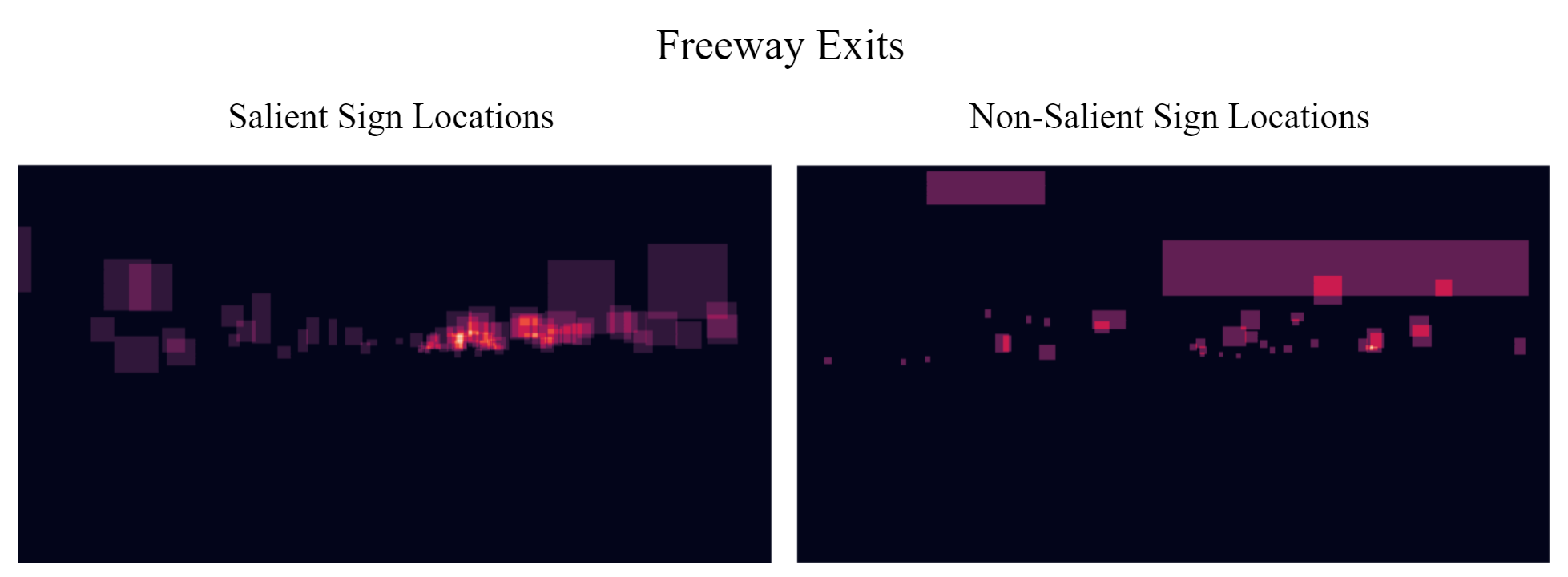}
         \caption{}
        \end{subfigure}
         \begin{subfigure}[b]{.49\textwidth}
         \centering
         \includegraphics[width=\textwidth]{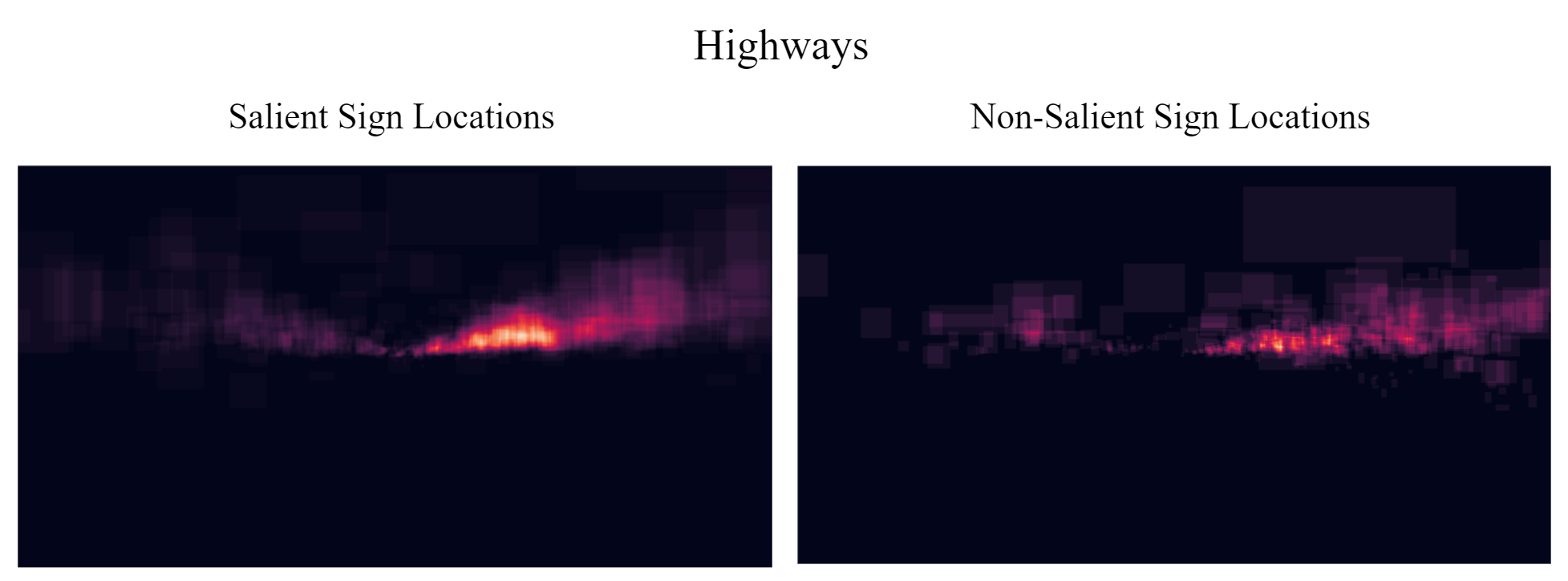}
         \caption{}
        \end{subfigure}
         \begin{subfigure}[b]{.49\textwidth}
         \centering
         \includegraphics[width=\textwidth]{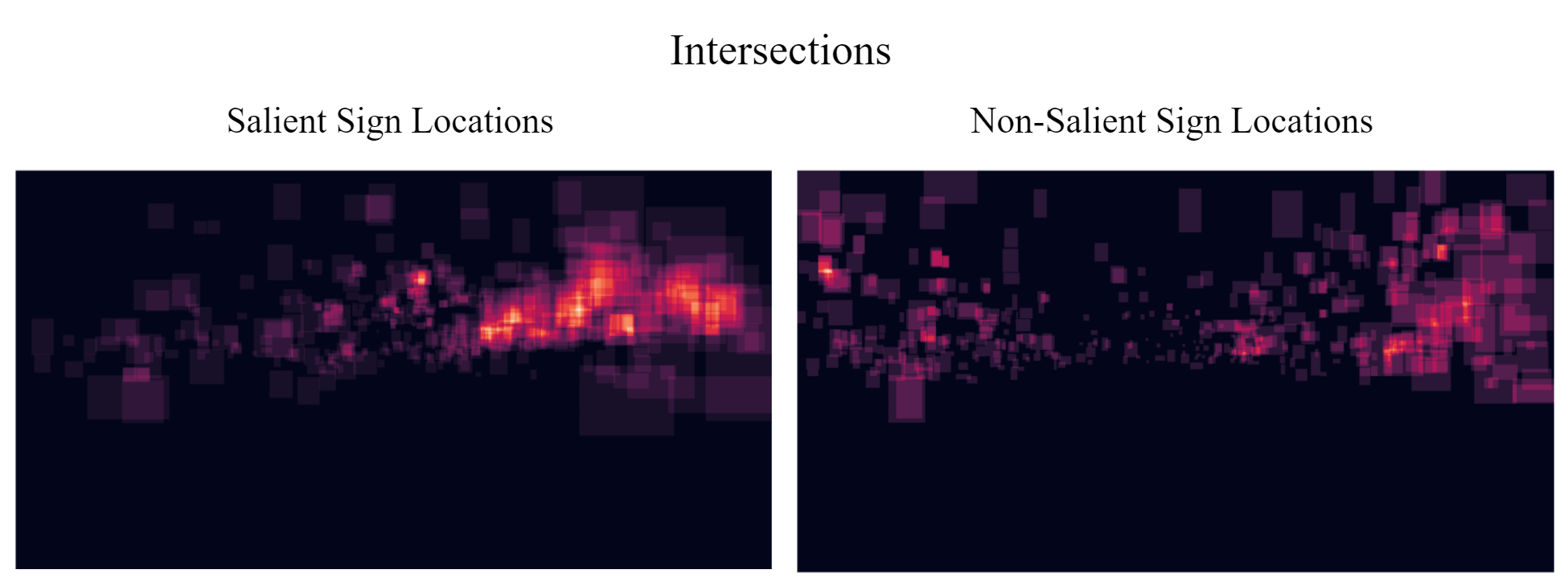}
         \caption{}
        \end{subfigure}
         \begin{subfigure}[b]{.49\textwidth}
         \centering
         \includegraphics[width=\textwidth]{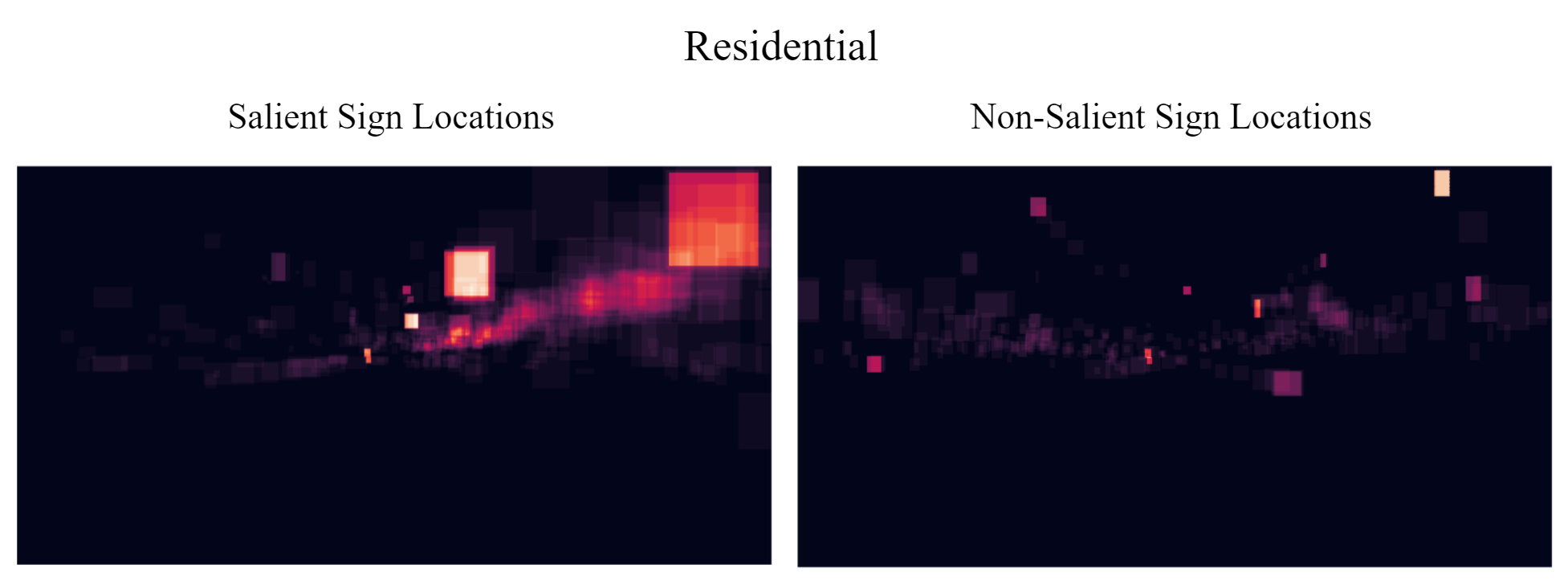}
         \caption{}
        \end{subfigure}
         \begin{subfigure}[b]{.49\textwidth}
         \centering
         \includegraphics[width=\textwidth]{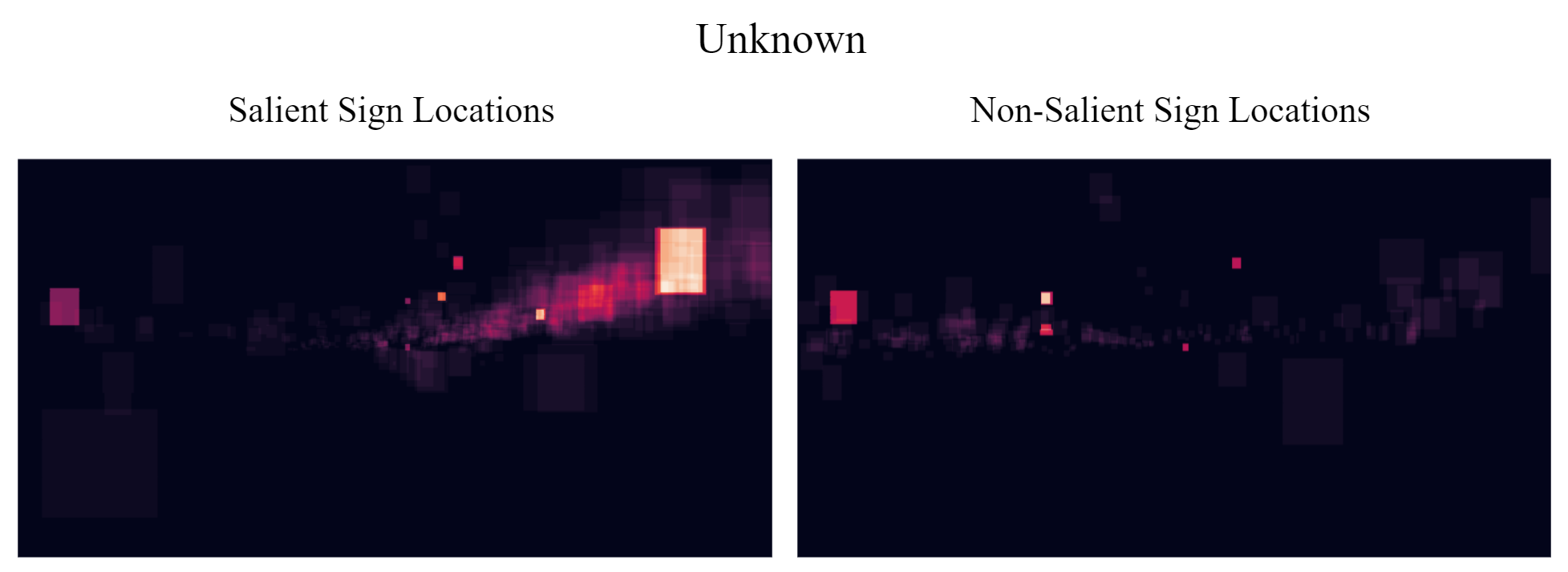}
         \caption{}
        \end{subfigure}

    \caption{Heatmaps illustrating frequency with which a pixel is occupied by a salient sign (left) or non-salient sign (right) for city streets (a), construction zones (b), freeway entrances (c), freeway exits (d), highways (e), intersections (f), residential (g), and unknown (h).}
    \label{fig:isfrontroadtype}
\end{figure*}

We further note that there is a relationship between expected sign location and road type, as illustrated by the heatmaps in Figure \ref{fig:isfrontroadtype}. This motivates experiments with a combined model, in which both road type and image coordinate features augment the convolutional output before the fully-connected layers. 

\subsection{Maneuver Augmentation}

A vehicle's intended motion contains information about which signs will be relevant. For example, if a vehicle is planning a right turn, it will likely be in a right lane, and a sign which reads ``Right Lane Must Turn Right" would be salient. Models which generate control for autonomous vehicles are of course unaware of the future trajectory, but it is reasonable that such model uses a series of planned maneuvers to navigate toward its goal. Accordingly, though the LAVA dataset does contain specific trajectory information, we use the coarse maneuver classification instead, as this is a reasonable substitute for the vehicle's intended path without assuming a particular trajectory. 

As in the previous augmentations, we integrate this classification into a one-hot encoded vector, appended to the feature set just before the fully-connected layers. We perform experiments in combining image, maneuver, road type, and image coordinate features, summarized in Table \ref{tableresults}. Convolutional methods and augmentations outperform the trivial classifier, achieving approximately 10\% improvement. Image coordinates (which indicate the location and distance of the sign relative to the ego vehicle position) do not appear to enhance beyond the ResNet50 baseline, though we expect that as the dataset grows, the performance of these models will improve as more examples of possible sign positions are used in training. From results on this dataset, augmentation with the vehicle's maneuver information shows the strongest results in determining sign salience. 

\begin{table}
\begin{center}
\begin{tabular}{||c c||} 
 \hline
 Model & Accuracy \\ 
 \hline\hline
 On-Right Baseline & 0.6650 \\
 \hline 
 ResNet50 & 0.7422 \\
 \hline
 Maneuver Augmentation & \textbf{0.7599} \\
 \hline
 Road Type Augmentation & 0.7153  \\
 \hline
 Coordinate Augmentation & 0.7231 \\
 \hline

 Coordinate \& Road Type & 0.7188 \\
 \hline
  Coordinate \& Maneuver & 0.7252 \\
  \hline
  Coordinate \& Road Type \& Maneuver & 0.7358  \\  
  \hline
\end{tabular}
\caption{Classification accuracy of the sign salience models.}
\label{tableresults}
\end{center}
\end{table}

\section{Conclusion \& Future Research}

In this work, we presented 
\begin{itemize}
    \item an analytical dimension of sign salience to weigh importance of particular traffic signs in path planning
    \item a traffic sign dataset which contains information on this property, with ability to infer road type and maneuver intent, and
    \item analysis of models for prediction of sign salience.
\end{itemize} 
The property of sign salience is intended for use in downstream path planning, where it could strategically penalize missed sign detections, sign classifications, and control decisions in salience-aware models. Extensions of the work include the conversion of salience from a binary to scalar property, and methods of determining scalar salience using subjective labelling between drivers. Sign salience is of further importance to driver-assistance systems which seek to understand human readiness and attention \cite{9564434}, and systems with augment a driver's scene awareness to the full surround \cite{rangesh2019no}. The LAVA dataset continues to grow, with an expected volume for future work which is four times the size available to this study. 

\section{Acknowledgements}
We appreciate the collaboration and support of our LISA sponsors, especially AWS Machine Learning Solutions Lab, and LISA colleagues. 

{\small
\bibliographystyle{ieee_fullname}
\bibliography{egbib}
}

\end{document}